\documentclass[conference]{IEEEtran}
\IEEEoverridecommandlockouts
\usepackage{cite}
\usepackage{amsmath,amssymb,amsfonts}
\usepackage{algorithmic}
\usepackage{graphicx}
\usepackage{textcomp}
\usepackage{xcolor}
\usepackage{booktabs} 
\usepackage{multirow} 
\usepackage{float}
\usepackage{lipsum} 
\usepackage{yfonts}
\usepackage{amsmath}
\usepackage{amsfonts}
\usepackage{amssymb}
\usepackage{graphicx}
\usepackage{algorithm}
\usepackage{algorithmic}
\usepackage{enumerate}
\usepackage{bm}
\usepackage{amsmath,amsthm,amssymb}
\usepackage{comment}
\usepackage{enumitem}
\usepackage{soul}

\newtheorem{mydefinition}{Definition}
\newtheorem{myremark}{Remark}
\newtheorem{myexample}{Example}
\floatname{algorithm}{Procedure}
\def\BibTeX{{\rm B\kern-.05em{\sc i\kern-.025em b}\kern-.08em
    T\kern-.1667em\lower.7ex\hbox{E}\kern-.125emX}}
    
\begin{document}

\title{An Empirical Study of the Role of Incompleteness and Ambiguity in
Interactions with Large Language Models}
\author{\IEEEauthorblockN{1\textsuperscript{st} Riya Naik}
\IEEEauthorblockA{
\textit{BITS Pilani, K K Birla Goa Campus}\\
Goa, India \\
p20210056@goa.bits-pilani.ac.in}
\\
\IEEEauthorblockN{2\textsuperscript{nd} Ashwin Srinivasan}
\IEEEauthorblockA{
\textit{BITS Pilani, K K Birla Goa Campus}\\
Goa, India \\
ashwin@goa.bits-pilani.ac.in}
\and
\IEEEauthorblockN{4\textsuperscript{th} Estrid He}
\IEEEauthorblockA{\textit{RMIT University}\\
Melbourne, Australia \\
estrid.he@rmit.edu.au}
\\
\IEEEauthorblockN{3\textsuperscript{rd} Swati Agarwal}
\IEEEauthorblockA{\textit{PandaByte Innovations Pvt Ltd}\\
India \\
agrswati@ieee.org}
}

\maketitle

\begin{abstract}
Natural language as a medium for human-computer interaction has long
been anticipated, has been undergoing a sea-change
with the advent of Large Language Models (LLMs) with 
startling capacities for processing and generating language. Many of us now treat LLMs as
modern-day oracles,  asking it almost any kind of question. Unlike its Delphic predecessor, consulting an LLM does
not have to be a single-turn activity (ask a question, receive an answer, leave); and---also unlike
the Pythia--it is widely acknowledged that answers from LLMs can be improved with additional context. In this paper, we aim to study when we need multi-turn interactions with LLMs to successfully get a question answered; or conclude that a question is unanswerable. We present a neural symbolic framework that models the interactions between human and LLM agents. Through the proposed framework, we define incompleteness and ambiguity in the questions as properties deducible from the messages exchanged in the interaction, and
provide results from benchmark problems, in which the answer-correctness is shown to
depend on whether or not questions demonstrate the presence of incompleteness or ambiguity
(according to the properties we identify).
Our results show multi-turn interactions are usually required for datasets which
have a high proportion of incompleteness or ambiguous questions; and that   
that increasing interaction length has the effect of reducing incompleteness
or ambiguity. The results also suggest that our measures of incompleteness
and ambiguity can be useful tools for characterising interactions with an LLM on question-answering
problems.
\end{abstract}

\begin{IEEEkeywords}
Information Systems, Question Answering, Query Analysis, Discourse, Dialogue, Pragmatics, and Large Language Models.
\end{IEEEkeywords}

\section{Introduction}
Imagine this conversation taking place in 1575. Pope Gregory XIII and the physician Luigi Lilio 
are discussing dates for Easter:

{\small{
\begin{tabbing}
[LL]: \= \kill
\textgoth{G}:\> Tell me, Luigi, in your calculation, will next year be a \\
\> leap year? \\
LL: \> Yes Your Holiness, since is divisible in four equal parts. \\
\textgoth{G}: \> I see. But then, 1500 would have been such a year. \\
LL: \> No Your Holiness. There is a correction made every century. \\
\textgoth{G}: \> Good. I assume that the same correction will be applied\\
\> in 1600, and it will not be a bisexstile year? \\
LL: \>  (apologetically) No, Holy Father. There is a further correction \\
\> once every  400 years, and  1600 will be a bissextile year.
\end{tabbing}
}}
\noindent
We now know that if the Holy Father had gone on to ask about 4000 A.D., the good doctor would have had to
admit to a further level of exception. The point here is not about the accuracy of the Gregorian calendar, but
that if the topic of conversation has multiple caveats, then it is not unusual for a back-and-forth
dialogue. The Abbot and Costello classic baseball Q\&A makes the same point when ambiguity is involved.
In both cases, it would be somewhat artificial if the questioner could only answer 1 question; or if the answerer provided
all that he knew about the topic in answer to the first question.

Conversational AI systems have long been anticipated, both in fiction and real life.
While the focus in fiction has inevitably been on the
modality of delivery (voice), the content has not been quite as demanding
(``Tea. Earl Grey. Hot.''), perhaps reflecting the state-of-the-art of natural language
processing at the time of conception. However, this latter aspect has recently
been undergoing rapid and repeated re-evaluation with the advent of Large Language
Models (LLMs). LLM-based technology is increasingly part of the design of systems involving some
form of human interaction. These systems control how we access information, interact with support services, and how we use devices. Their role is mainly aim to simulate human-like conversations, providing users with intuitive and effective responses \cite{kulkarni2019conversational}, often in the form of answers 
to questions. It is therefore not unexpected that a significant amount of academic and industrial
research and development is invested into arriving at a better understanding
of interactions with LLMs. In this paper, we seek to contribute to this understanding by examining
two aspects of interactions with LLMs, namely, \textit{single-turn} and \textit{multi-turn} interactions.

It is now evident that modern LLMs can store very large amounts of context-information accumulated during the course of an interaction (input token-lengths are now millions of tokens, and only expected to increase \cite{achiam2023gpt}). It is also known that techniques such as in-context learning that provide feedback in the form of clarificatory
instances and domain knowledge can greatly assist an LLM in retrieving relevant answers to queries \cite{dong2024survey,chitale2024empirical}. It is also the case that the amount of such clarificatory information can often be surprisingly small \cite{li2025review}. It is, therefore, the case that LLMs have the mechanisms necessary for conducting interactive conversations \cite{daryanto2025conversate}. But, a question of some conceptual importance is this: when do LLMs need interactive feedback?
There are 3 ways to approach such a question. First, from a mathematical point-of-view, by assessing if there
is an information-theoretic gap in what is known and what is asked. The LLM may then need clarification to
reduce this uncertainty (used in a mathematical sense). Secondly, is a technological approach. In this, we
could attempt to see if current LLM technology (software or hardware) make it impossible for the LLM to distinguish
between some aspects of the problem (for example, perhaps positional encoding makes it impossible to see the
question is about sets and not sequences). Finally, there is a behavioral approach, in which we
try to assess the conditions for feedback by examining the sequence of messages sent between
the participants in an interaction. It is this third category that is of interest to us in this paper.
Specifically, we focus on question-answering tasks, and attempt to identify conditions from the
message-sequence of:
\begin{enumerate}
    \item Whether a question is incomplete, requiring further clarification on what is meant by specific terms in the response; and
    \item Whether a question is ambiguous, requiring further clarification of the conditions under which the response is sought
\end{enumerate}

The rest of the paper is organized as follows. In Section \ref{sec:interact} we
provide an interaction model for an interactive question-answering system.
We use this model as the basis for specifying the presence of incompleteness and ambiguities.
In Section \ref{sec:expt} we describe an empirical study to assess the occurrence of incompleteness or ambiguities or both than single-turn interactions.
Section \ref{sec:relwork} summarises some related work, Section \ref{sec:concl} concludes the paper, and Appendix \ref{sec:appendix} contains examples extracted from the messages exchanged between the human agent and the model, illustrating incomplete and ambiguous questions.

\section{A Simple Messaging System  for Interaction}
\label{sec:interact}

In this section, we describe a simple message-based
communication mechanism between a pair of agents.
Communication between agents consists of messages from
a sending agent to a receiving agent.

\begin{mydefinition}[Messages]
A message  is a 3-tuple of the form $(a,m,b)$ where $a$
is the identifier of the sender; $b$ is the identifier of the
receiver; and $m$ is a finite length message-string. 
\end{mydefinition}

\noindent
We define the following categories of 
message-strings:

\begin{description}
\item[Termination.] The message-string  $m = ''\Box''$
	denotes that sender is terminating the communication
	with receiver.
\item[Question.] The message-string $m = ''?_n(s)''$ denotes
	that the sender is sending a question $s$ with identifier
	$n$ to receiver.
\item[Answer.] The message-string $m = ''!_n(s)''$ denotes that the
	sender is sending an answer $s$ with identifier
	$n$ to receiver.
\item[Statement.] The message-string $m = ''\top(s)''$ denotes that
	sender is sending a statement to receiver.
\end{description}

\noindent
In the rest of the paper, we omit the identifier n as long as the context is clear. In the above definitions, s represents a sequence of message strings. However,
in this paper, in $?(s)$ $|s| = 1$ (that is, only 1 question is allowed at a time).
There can be multiple answers or even no answers for a question. That is,
in $!(s)$, $|s| \geq 0$. Additionally, in $\top(s)$, we will require $|s| \geq 1$.
If ordering is unimportant, we will sometimes show $s$ as a set instead of a sequence.
In all cases, if $|s| = 1$, we will simply denote the message by the singleton element,
dispensing with the sequence (or set) notation.

\noindent
The interaction between a pair of agents consists of 1 or more turns.

\begin{mydefinition}[Interaction]
A (1-step) turn from agent $a$ to agent 
$b$ is the pair of messages $ (M_1,M_2)$, where
$M_1 = (a,m_1,b)$, $M_2 = (b,m_2,a)$, and $m_1 \neq \Box$.
A k-step turn is the sequence $\langle T_1,T_2,\cdots,T_k \rangle$,
where $T_i$ $(1 \leq i \leq k)$ is a 1-step turn from $a$ to $b$.
Similarly for 1-step and $k$-step turns from $b$ to $a$.
We will call a sequence of 1 or more turns between $a$ and
$b$ an interaction between $a$ and $b$.
\end{mydefinition}

\begin{myexample}[Interaction]\label{ex:hm1}
Let $h$ denote a human agent and $m$ denote a machine.
Below is an example of a possible 3-step turn interaction between $h$ and $m$: \\
{
$\langle$ 
$(h,\top(\mathrm{Child~x~has~a~height~is~4~ft.}),m)$,$(m,\top(\mathrm{ok}),h)$,\\
$(h,\top(\mathrm{The~height~of~child~y~is~the~square~root~of~the}$ \\ $\mathrm{~height~of~child~x}),m)$,
$(m,\top((\mathrm{ok}),h)$,\\
$(h,?_1(\mathrm{What~is~the~height~of~y}),m)$, $(m,!_1(\mathrm{y~is~+2~or~-2}),h)$
$\rangle$
}
\end{myexample}

\begin{myexample}[Interaction (contd.)]\label{ex:hm2}
A 4-step turn interaction between $h$ and $m$ is:\\
$\langle$
$(h,\top(\mathrm{Child~x~has~a~height}$ $\mathrm{is~4~ft.}),m)$,
$(m,\top(\mathrm{ok}),h)$,\\
$(h,\top(\mathrm{The~height~of~child~y~is~the~square~root~of~the}$ \\$\mathrm{~height~of~child~x}),m)$,
$(m,\top(\mathrm{ok}),h)$,\\
$(h,?_1(\mathrm{What~is~the~height~of~y}),m)$,
$(m,!_1(\mathrm{y~is~+2~or~-2}),h)$,\\
$(h,\top(\mathrm{Your~answer~is~not~completely~correct~since~height}$ \\$\mathrm{~has~to~be~positive}),m)$,
$(m,!_1(\mathrm{y~is~+2}),h)$
$\rangle$
\end{myexample}

We note that each turn consists of a sequence of 2 messages.
Thus, with every interaction consisting of $k$ turns
$\langle T_1,\cdots,T_k\rangle$ there exists a corresponding
sequence $\langle M_1,M_2,\cdots,M_{2k-1},M_{2k} \rangle$ messages
and $\langle m_1,m_2,\cdots,m_{2k-1},m_{2k} \rangle$ message-strings.
By construction the sequence
$\langle m_1, m_3,\ldots,m_{2k-1}\rangle$ will be from agent $a$
to agent $b$, and
$\langle m_2,m_4,\ldots,$ $m_{2k}\rangle$ will be from agent $b$
We denote these as $\langle m_{ab} \rangle$ and
$\langle m_{ba} \rangle$ for short.  

\noindent
Interaction sequences allow us to define the \textit{context} for
an agent. We assume any agent has access to a (possibly empty)
set of prior statements, which we call \textit{background knowledge}.

\begin{mydefinition}[Context]
\label{def:context}
Let $a$ and $b$ be agents with background knowledge $B_a$ and $B_b$ respectively, prior to any interaction.
Without loss of generality, let $(T_1,T_2,\cdots,T_k)$ be a k-step interaction
from $a$ to $b$. We denote the context
for $a$ on the the $i^\text{th}$
turn $T_i$ as $C_{a,i} = $
$B_a \cup \{m_1,m_2,\cdots,m_{2i-2)}\}$, and the context for $b$ on the $i^\text{th}$ turn
$C_{b,i}$ $=$
$B_b \cup \{m_1,m_2,\cdots,m_{2i-1}\}$.
\end{mydefinition}

\noindent
In this paper, we are interested in question-answer sequences
occurring in an interaction. These are obtainable by examining
the messages exchanged.

\begin{mydefinition}[Questions and Answers]
\label{def:qanda}
Let $(T_1, \cdots, T_k)$ be a k-step interaction between
$a$ and $b$, and $(m_1, m_2,\cdots, m_{2k-1},m_{2k})$
be the corresponding message-strings. Let $m_{ab}$
and $m_{ba}$ be the message-string sequences from
$a$ to be $b$ and vice-versa. Let
${QA}_{ab}$ be
the sequence $((q_1,a_1),\cdots, (q_j,a_j))$
s.t.:
(1) for every $(q_i,a_i)$ in ${QA}_{ab}$,
    $?_{\alpha_i}(q_i) \in m_ {ab}$; and
(2) $a_i = \cup ~!_{\alpha_i}(s)$, where $!_{\alpha_i}(s) \in  m_{ba}$.
We will call ${QA}_{ab}$ the
question-answer sequence for the interaction between $a$ and $b$.
Similarly for a question-answer sequence from $b$ to $a$.
\end{mydefinition}

\noindent
It is sometimes helpful to identify the set of questions sent
by $a$ to $b$, or $Q_{ab}$ as
= $\{Q: (Q,\cdot) in \langle {QA}_{ab} \rangle \}$. A similar
set of questions from $b$ to $a$ can also be identified.

\subsection{The Oracle}
\label{sec:oracle}

We define a special agent $\Delta$ called the
{\em oracle\/}. The oracle's answers are taken to be always correct.
The oracle is assumed to know everything.

\begin{myremark}[Interaction with the Oracle]
We note the following special features of the oracle:
\begin{itemize}
    \item $\Delta$ knows everything upto the    
        present, including the
        content of conversations between any non-oracular agents;
    \item Only a 1-step interaction is allowed between a non-oracular
        agent $a$ and $\Delta$. The interaction consists of a turn $T$ where:
        \begin{itemize}
            \item[$\rightarrow$] $T = ((a,?q,\Delta),(\Delta,!(s),a))$; or 
            \item[$\rightarrow$] $T = ((a,?q,\Delta),(\Delta,\Box,a))$.
        \end{itemize}
    \item The answer(s) provided by $\Delta$ are always correct.
\end{itemize}
\end{myremark}


\begin{myexample}[Interaction with an Oracle]\label{ex:ho1}
We consider the example again, this time with the 
human agent interacting with an oracle $\Delta$.
A 1-step interaction between $h$ and $\Delta$ is:\\
$\langle$
$((h,?_1(\mathrm{What~is~the~height}$ $\mathrm{of~child~y}),\Delta)$,\\
$(\Delta,!_1((\mathrm{y=+2},\Box)),h))$
$\rangle$
\end{myexample}

\noindent
The oracle allows us to categorise questions and answers.

\begin{mydefinition}[Incomplete Question]
Without loss of generality, let $(T_1,T_2,\cdots,T_k)$ be a k-step interaction
from $a$ to $b$. Let $C_{b,i}$ denote the context for $b$ on the the $i^\text{th}$ turn. Let $T_i = ((a,?(q),b),\cdot)$, where agent a sends question q to b.
Let $((b,?(q),\Delta),(\Delta,!(s),b))$
be an interaction between $b$ and $\Delta$. If $s=\Box$, we conclude that  $q$ is incomplete.
In such a case, we will also say it is incomplete for $b$ given
$C_{b,i}$.
\end{mydefinition}

\noindent
That is, a question is incomplete, if the oracle does not give an answer. This is because
if the oracle cannot answer, neither can $b$. Similarly:

\begin{mydefinition}[Ambiguous Question]
Without loss of generality, let $(T_1,T_2,\cdots,T_k)$ be a k-step interaction
from $a$ to $b$. Let $C_{b,i}$ denote the context 
for $b$ on the $i^\text{th}$ turn. Let $T_i = ((a,?(q),b),\cdot)$, where agent a sends question q to b.
Let $((b,?(q),\Delta),(\Delta,!(s),b))$
be an interaction between $b$ and $\Delta$. If $|s| > 1$ then we will say $q$ is ambiguous.
In such a case, we will also say $q$ is ambiguous for $b$ given $C_{b,i}$.
\end{mydefinition}

\noindent
That is, a question is ambiguous if the oracle returns more than one answer.
We assume that questions that are either incomplete or ambiguous, but not both at once.
We are still left with the impractical requirement of needing to consult the oracle
to decide whether a question is one or the other. We propose the following tests to
detect the possible presence of incompleteness and ambiguity using just the
interaction sequence between non-oracular agents.

\begin{mydefinition}[Possibly Incomplete Question]
\label{def:pi}
Let $I = (T_1,T_2,\cdots,T_k)$ be a k-step interaction between $a$ and $b$.
Let  $T_i = ((a,?_\alpha(q),b),(b,?_\beta(s1),a))$; and
$T_{i+1} = (a,!_\beta(s2),b),(b,s3,a))$, where $s3$ can be any statement. Then
we will say $q$ is a possibly incomplete question.
\end{mydefinition} 

\begin{mydefinition}[Possibly Ambiguous Question]
\label{def:pa}
Let
 $I = (T_1,T_2,\cdots,T_k)$ be a k-step interaction between $a$ and $b$.
Let $T_i = ((a,?_\alpha(q),b),(b,!_\alpha(s1),a))$; and
$T_{i+1} = (a,\top(s2),b),(b,s3,a))$. Then
we will say $q$ is a possibly ambiguous question.
\end{mydefinition} 

\begin{myremark}
We note the following about these definitions:
\begin{itemize}
    \item We note that the definitions for detecting possible incompleteness and
    possible ambiguity require interactions of more than 1 step, i.e., multi-turn interactions. The definitions
    can be generalised to span more than 2 turns, but the existence of 
    at least 2 turns demonstrating the pattern shown is taken to be sufficient here;
\item It is possible that $b$ may send $a$ questions to clarify further. This may
    result in that subsequent question itself being incomplete or ambiguous. In this
    paper, we will be concerned only with the initial question from $a$ to $b$.
\item The definitions are axiomatic in nature. That is, we will be investigating
    (empirical) consequences assuming the definitions hold.
\item It is possible that by examining the sequence of messages and collating the clarificatory
    statements and answers from $a$, we can re-organise the interaction such that $a$ provides
    $b$ with all the relevant information before asking a question. This may
    result in a single-turn interaction, provided the information from
    $a$ is sufficient to answer the question correctly, and $b$ at least as rational as the oracle
\end{itemize}
\end{myremark}
The oracle can also be seen as a source of ``ground-truth'', thus providing
us with a way of deciding correctness of answers, and consequently assessing
the quality of answers in an interaction between non-oracular agents $a$ and $b$.


\begin{myexample}[Answer Quality]
From Example \ref{ex:hm1},
${QA}_{ab}$ = \\
$\langle$
$(?_1(\mathrm{What~is~the~height}$ $\mathrm{of~y}), \{!_1(\mathrm{y~is}$ $\mathrm{+2~or~-2})\})$
$\rangle$\\
From Example \ref{ex:hm2},
${QA}_{ab}$ =
$\langle$
$(?_1(\mathrm{What~is~the~height}$ $\mathrm{of~y}),\{!_1(\mathrm{y~is}$ $\mathrm{+2~or~-2}), !_1(\mathrm{y~is~+2})\})$
$\rangle$\\
From Example \ref{ex:ho1},
${QA}_{a\Delta}$ =
$\langle$
$(?_1(\mathrm {What~is~the~height}$ $\mathrm{of~y}), \{!_1(\mathrm{y~is~+2})\})$
$\rangle$

\noindent
Then, the answer quality of the interaction  Example \ref{ex:hm1} is
a function of:
$\{$
$(\{\mathrm{y~is}$ $\mathrm{+2~or~-2}\}$ $\{\mathrm{y~is~+2}\})$
$\}$; the answer quality of the interaction Example \ref{ex:hm2} is
a function of:
$\{$
$(\{\mathrm{y~is~+2}$ $\mathrm{or~-2}\},\{\mathrm{y~is}$ $\mathrm{+2}\})$
$(\{\mathrm{y~is~+2}\},\{\mathrm{y~is~+2}\})$
$\}$
 \end{myexample}

 \section{Empirical Evaluation}
 \label{sec:expt}

We are concerned with human-LLM interaction on question-answering problems. Using some prominent benchmark
datasets, our experimental goals are twofold:
\begin{itemize}
    \item We want to determine if current LLM implementations are sufficiently
        powerful to answer most questions correctly with single-turn interactions;
    \item When LLM implementations require multi-turn interactions, we want to
        identify the extent to which these are due to the presence of possible
        incompleteness or ambiguities in the questions.
\end{itemize}

\subsection{Materials}
\label{sec:mat}

\subsubsection*{Datasets}
In our empirical study, we evaluate the QA systems on six datasets with different characteristics. First, NQ-open (Natural Questions open), which is a large-scale benchmark, featuring open domain real user queries and answers annotated from Wikipedia articles \cite{kwiatkowski2019natural}.
Second, SQuAD (Stanford Question Answering Dataset), which is a widely used dataset for machine reading comprehension, consisting of over 100,000 questions based on Wikipedia articles\cite{rajpurkar2018know}.
Third, MedDialog covering 0.26 million conversations between patients and doctors curated to understand real-world medical queries \cite{zeng2020meddialog}.
Fourth, MultiWOZ (Multi-domain Wizard-of-Oz), a dataset covering multiple domains such as hotels, restaurants, and taxis with 8438 task-oriented dialogues \cite{budzianowski2018multiwoz}.
Fifth, ShARC (Shaping Answers with Rules through Conversation), a multi-turn dataset that focuses on 32k task-oriented conversations with reasoning covering multiple domains\cite{saeidi2018interpretation}.
Sixth, the AmbigNQ dataset designed to handle ambiguous questions focusing on event and entity references covering 14,042 NQ-open questions\cite{min2020ambigqa,kwiatkowski2019natural}.
The second, fourth, and fifth datasets include questions along with the relevant context to provide answers, while the first, third, and sixth datasets consist of question-answer pairs. In the case of MultiWOZ and MedDialog, the human answering agent is replaced with LLM.

\subsubsection*{Algorithms and Machines}
We use the following Large Language Model and composable framework: (a) GPT3.5-Turbo: LLM is used to retrieve the answer to the question in single and multiple-turn interaction settings. (b) LangChain: the framework is used to query GPT3.5-Turbo. All implementations are in Python 3.10, with API calls to the respective model engine.

Our experiments are conducted on a workstation based on Linux (Ubuntu 22.04) with 256GB of RAM, an Intel i9 processor, and an NVIDIA A5000 graphics processor with 24GB memory. 

\subsection{Method}
\label{sec:meth}
The method adopted follows the steps given below:

\begin{enumerate}[label=\arabic{enumi}.]
    \item \textbf{Initialization:}  
    Let \( h \) represent a human agent and \( \lambda \) denote the LLM (Language Model).
    
    \item \textbf{Processing:}  
    For each dataset \( d \):
    \begin{enumerate}
        \item Initialize \( I_d = A_d = \emptyset \) as empty sets.
        \item For each data point \( x = (q, a) \) in \( d \):
        \begin{enumerate}
            \item Simulate an interaction \( S_x \) starting with the first message as \( (h, ?(q), \lambda) \).
            \item Update the interaction category of \( S_x \) based on predefined patterns:
            \begin{itemize}
                \item[] If \( S_x \) matches the pattern defined in Definition~\ref{def:pi}, then \( I_d := I_d \cup \{q\} \); Else if \( S_x \) matches the pattern defined in Definition~\ref{def:pa}, then \( A_d := A_d \cup \{q\} \).
            \end{itemize}
        \end{enumerate}
        \item Compute the proportions for the dataset \( d \):
        \[
        PI_d = \frac{|I_d|}{|d|}, \quad PA_d = \frac{|A_d|}{|d|}
        \]
    \end{enumerate}
    
    \item \textbf{Comparison:}  
    Examine the relationship of proportions \( PI_d \) and \( PA_d \) to the proportion of interactions that were correctly answered in a single-turn.
\end{enumerate}

The following additional details are relevant:
\begin{itemize}
    \item We make API calls to GPT-3.5-Turbo with the temperature set to 0.7 as it provides a balanced trade-off between creativity and reliability while generating text \cite{peeperkorn2024temperature}.
    \item Using the question and context, we assemble a prompt with instructions. The LLM uses this prompt to generate responses.
    \item We, as a human agent, respond with clarification to improve the quality and accuracy of the final answer.
    \item Quantitative performance comparison is performed as follows: the proportions \( PI_d \) and \( PA_d \) are compared to assess the alignment of the model's responses with predefined patterns. This comparison provides insights into the distribution of interactions across different categories that seek multiple interactions
\end{itemize}
Since incomplete and ambiguous questions are only properties defined on interactions with at least 2 turns,
${PA}_d = {PI}_d$ will be $0$ if all questions can be answered correctly in 1 turn. For questions requiring
longer interactions, the comparison step above will allow us to estimate
the proportion of multi-turn interactions in which the question initiating the interaction is either
incomplete or ambiguous (to the extent defined by Defns. \ref{def:pi} and \ref{def:pa}).

\subsection{Results}
\label{sec:res}

The principal results from the experiments are in Table \ref{tab:dataset_analysis}. The
results have been grouped into 3 categories: (C1) Datasets for which correct answers are
largely obtained in a single-turn; (C2) Datasets which require multi-turn interactions, in which the questions posed by the human-agent
are mainly incomplete (according to Defn. \ref{def:pi}); (C3) Datasets which require multi-turn interactions, in which the
questions posed by the human agent are mainly ambiguous (according to Defn. \ref{def:pa}).

It is evident from the tabulations that: (i) It is not the case that the LLMs are
able to resolve questions in all datasets with a single-turn; and (ii) For those datasets requiring
multiple turns (that is, datasets in (C2) and (C3)),  incomplete and ambiguous questions
(taken together) match the proportion of questions incorrectly classified in a single turn.
That is, all questions requiring more than 1 turn to answer correctly are either incomplete
or ambiguous (according to Defns \ref{def:pi} and \ref{def:pa}).

\begin{table}[h!]
\centering
\caption{Proportions of incomplete and ambiguous questions. The last column is proportion of
    interactions in which a correct answer is obtained from the LLM in a single-turn. The
    proportion incorrectly answered in 1 turn is therefore 1 minus the entry in this last column.}
\begin{tabular}{@{}lcccccc@{}}
\toprule
                 &                   & \multicolumn{2}{c}{\bf{Questions}} & \bf{Answers} \\
\textbf{Category} & \textbf{Dataset} & \textbf{Incomplete} & \textbf{Ambiguous} & \textbf{Correct} \\

                  &    $(d)$         &    $({PI}_d)$       & $({PA}_d)$         &  {(after 1 turn)} \\ \midrule
C1  & NQ-open & 0.02 & 0.17 & 0.81\\
& SQuAD & 0.00 & 0.08 & 0.92 \\[6pt]
    
C2 & MedDialog  & 0.92 & 0.08 & 0.00 \\[6pt]

C3 & MultiWOZ  & 0.21 & 0.75 & 0.04 \\
& ShARC & 0.28 & 0.61 & 0.11  \\ 
& AmbigNQ & 0.01 & 0.36 & 0.63   \\ \bottomrule
\end{tabular}
\label{tab:dataset_analysis}
\end{table}

\noindent
We now examine these results in more detail:

\noindent
{\bf Datasets.} Recall that both SQuAD, NQ-open, and AmbigNQ are datasets for which answers are largely fact-based.
    The results suggest that for such kinds of Q\&A datasets, LLMs are able to identify correct answers
    largely in a single-turn. This is not surprising, especially for SQuAD and NQ-open, which has neither
    a high proportion of incomplete or ambiguous questions.  We note the AmbigNQ dataset
    was devised as a modification of the NQ-open dataset, by introducing ambiguity in the
    questions. This is supported here, with our categorisation confirming AmbigNQ as having
    a significantly higher proportion of Ambiguous questions. However, this proportion is
    modest, suggesting that the background information available to the LLM is able to
    resolve the ambiguity in the questions (the AmbigNQ dataset was not intended to test
    LLM performance). Both SQuAD and NQ-open were
    devised to be tests of single-turn performance, with correct answers expected to be largely
    obtained in 1 turn; and less likely to be possible in 1 turn with AmbigNQ. This
    is consistent with our results, although, by our definitions, MedDialog would appear to contain a far higher proportion of incomplete questions.

\noindent
{\bf Context.} Context (as defined by Def. \ref{def:context}) plays an important role in reducing incompleteness and
    ambiguity.
    Recall that as the length of interaction
    increases, the context available to the LLM increases. To simulate the effect of
    increasing
    on the incompleteness or ambiguity of the initial question, we progressively
    include the information from additional turns as part of the initial context,
    and determine the proportion of initial questions identified as incomplete or ambiguous.
    Table \ref{tab:morecontext} tabulates the results of this retroactive increase in
    context. It is evident that as information is included from additional turns, the proportion of incorrect (incomplete or ambiguous) questions decreases. 
    Questions that are no longer incorrect are able to be answered in a single turn with the additional context, which mirrors the increase tabulated in the last column. We provide an example of one such instance in the Appendix \ref{sec:appendix}. This increase in proportions is especially seen in the MedDialog, MultiWOZ, and ShARC datasets, which rely heavily on context information about the particular situation or individual. We witness a shift from incomplete to ambiguous in MedDialog, as a consequence of conversations involving subjective input and multiple interpretations even after the context is added. This complexity is less seen in the MultiWOZ and ShARC datasets, where task-specific structured goals minimize ambiguity.

\begin{table}[h!]
\centering
\caption{Role of increasing context on the proportions of incomplete and ambiguous interactions. context
    provided to the LLM increases as the number of turns ($k$) increases. The last column tabulates the
    proportion of correct answers after the corresponding number of turns.} 
\begin{tabular}{@{}lcccccc@{}}
\toprule
                 &  \textbf{Context from} & \multicolumn{2}{c}{\bf{Questions}} & \bf{Answers} \\
\textbf{Dataset} & \textbf{Turns} & \textbf{Incomplete} & \textbf{Ambiguous} & \textbf{Correct} \\ 
    $(d)$        & \textbf{(k)}   &  $({PI}_d)$         & $({PA}_d)$ &  {(after 1 turn)} \\ \midrule
NQ-open & 1 & 0.02 & 0.17 & 0.81            \\
      & 2 & 0.00 & 0.13 & 0.87       \\
      & 3 & 0.00 & 0.11 & 0.89  \\[6pt]
SQuAD & 1 & 0.00 & 0.08 & 0.92            \\
      & 2 & 0.00 & 0.05 & 0.95       \\
      & 3 & 0.00 & 0.03 & 0.97  \\[6pt]
MedDialog & 1 & 0.92 & 0.08 & 0.00 \\
          & 2 & 0.21 & 0.61 &  0.18\\
         & 3 & 0.18 & 0.56  & 0.26 \\[6pt]
MultiWOZ & 1 & 0.21 & 0.75 & 0.04      \\
       & 2   &  0.19    & 0.56     & 0.25 \\ 
       & 3   &   0.18   &   0.34   & 0.48 \\[6pt]
ShARC & 1 & 0.28 & 0.61 & 0.11       \\ 
      & 2 & 0.02 & 0.38 & 0.60       \\
      & 3 & 0.01 & 0.16 & 0.83      \\[6pt]
AmbigNQ & 1 & 0.01 & 0.36 & 0.63 \\
        & 2 & 0.00 & 0.31 & 0.69 \\
        & 3 & 0.00 & 0.22 & 0.78 \\
     \bottomrule
\end{tabular}
\label{tab:morecontext}
\end{table}

\section{Related Work}
\label{sec:relwork}

Conversational AI is gathering substantial attention recently, leading to significant advances in natural language processing (NLP) and dialog management\cite{fu2022learning}. Preliminary efforts in conversational systems focused on single-turn interactions; while these systems worked well for simple queries, they could not handle natural and engaging context in conversations \cite{na-etal-2022-insurance}. 
To improve this, deep learning models learned long context management with multi-turn interactions by maintaining a contextual memory across turns \cite{zhao2018multi}. Transformer models such as GPT \cite{radford2018improving} and BERT \cite{kenton2019bert} enabled systems to handle both single-turn interactions with precision and multi-turn interactions with coherence.

\noindent
{\bf Interactions.} Recent works in large language models (LLM) are now changing the story by revolutionizing single-turn performance \cite{huo2023retrieving,tan2023can}. As researchers are delving deeper into deducing interactions, the gap in understanding nuanced behavioral differences between single and multi-turn continues to grow. There are few works that examine the comparative performance of these interactions. For instance, Burggr et al.\cite{burggraf2022preferences} highlight user preferences for multi-turn interactions in an automotive context, suggesting that extended conversations enhanced user satisfaction. Similarly, Sorathiya et al.\cite{sorathiya2021multi} introduce methods to transform and utilize single-turn data for multi-turn conversations, enabling better performance and simulating real-life conversation in domains such as medicine \cite{li2023s2m}. Researchers provide a comprehensive review of the evolution of question-answering systems, with a growing emphasis on multi-turn approaches \cite{zaib2022conversational}. Furthermore, a recent study also details the implementation and experimental evaluation of a protocol for intelligible interaction between a human expert and an LLM \cite{srinivasan2024implementation}. This study enhances human-machine intelligibility by offering a protocol for structured interaction between humans and LLM. While these studies provide valuable insights, research on understanding human-machine interactions at the grassroots in the era of LLM remains under-explored. 

\noindent
{\bf Incompleteness and Ambiguity.} LLMs are designed to process and generate human-like text, but they can still misinterpret context and provide an answer that is inscrutable. Clarifactory feedback is essential for dialog to remain accurate and understandable. Identifying the presence of incompleteness and ambiguity in the question is imperative to include clarificatory feedback.
To understand incomplete questions, Addlesee and Damonte \cite{addlesee2023understanding} build pipelines that parse incomplete questions and repair them following human recovery strategies. They label underspecified queries with missing entities as incomplete questions. Similarly study by Kumar and Joshi \cite{kumar2017incomplete} considers an incomplete question as one which lacks either of the three: topic presentation, adjective expression, and the inquiring part. A series of techniques are employed to detect these ambiguity in the questions \cite{gao2021answering}. Graphical information is used to define the ambiguity of a query in terms of concepts covered and the similarities between the answer and the query \cite{banerjee2021detecting}. A few studies frame it as scope ambiguity when a sentence contains multiple operators and overlapping scope \cite{kamath2024scope}. Rasmussen \cite{rasmussen2022broad} expands on this by defining disambiguation using Roberta as the span classification task. Recent research uses LLMs to automate ambiguity detection by injecting ambiguous patterns extracted from a human-annotated set of relational tables. Researchers further quantify the level of ambiguity between questions and tabular data using customized performance metrics \cite{papicchio2024evaluating}. It is also defined by Cole et al. \cite{cole2023selectively} as the uncertainty of the question's intent. Our paper builds upon similar veins, aiming to bridge the gap and generalize the concept of incompleteness and ambiguity in natural language interactions.  
\section{Conclusion}
\label{sec:concl}

Our interest is in question-answering (QA) systems in natural language. While the
use of natural language has long been a desirable requirement for interfaces to
computers, the possibility of achieving it has proved significantly harder until the
advent and development of large language models (LLMs). Even at this early stage of
LLM technology, we are able to focus on aspects of specific issues that arise with
the use of natural language, within the ambit of QA systems. In this paper, we 
have concentrated on two such aspects: incompleteness and ambiguity. Both feature
during the course of an interaction in natural language, and we propose
ways of identifying the occurrence of one or the other, as properties defined on the syntax of
messages exchanged between any pair of agents in a question-answering system. The definitions
are not intended to be normative, but do have some intuitive features. Specifically, properties
of incompleteness or ambiguity in questions are only detectable in interactions that at least 2 turns and
our experiments with human-LLM interactions, across a range of benchmark QA datasets suggest that:
(a) For multi-turn interactions, the initial question posed fall into either being incomplete or
ambiguous; and
(b) As the number of turns increase, the proportion of correct answers from the LLM increases.
This latter finding in itself if not surprising, given that LLMs have shown the ability to adapt
quickly with increasing context. However, we can examine what would have happen, had the additional context been provided up-front. In that case,
we find the proportion of incomplete and ambiguous questions decreases, and the LLM is able to identify correctly
more answers in a single-turn. This indirectly supports our conjecture that decreasing incompleteness and
ambiguity  will lead to shorter, and more accurate responses from the LLM. To the best of our knowledge, this
is the first time formal definitions of these properties have been defined for human-LLM interactions and
tested empirically on a several diverse kinds of QA benchmarks.

There are a number of ways in which the paper here can be extended. First, on the experimental front, evidence
from additional datasets is always desirable. Also of  interest to see improvements in LLM technology
allow the internal resolution of incompleteness and ambiguity without requiring additional
input from the human. This amounts to performing some kind of self-checking and verification, wither by access
to additional information not within the LLMs knowledge-store (through some form of retrieval-augmented generalisation, for example \cite{mansurova2024qa}). It is also the case that we have provided experimental results only
for the initial question posed by the human. It is evident that the definitions we provide can equally be
used to identify incompleteness or ambiguity in subsequent questions in the interaction. Presumably, the
occurrence of such questions will increase the length of interaction, and even result in fluctuations in the
accuracy of response. It is of interest to understand these aspects of the interaction as well.
On the conceptual front, our definitions for incompleteness and ambiguity are based on mutually exclusive patterns.
Presumably questions can be both incomplete and ambiguous, and the definitions would need to be generalised to
accommodate this. We do not see any reason in principle why this cannot be done, and the result should allow
a finer-grained understanding of multi-turn interactions with an LLM.
\bibliographystyle{IEEEtran}
\bibliography{references}

\appendix
\section{Appendix A}
\label{sec:appendix} 
\noindent
Tables \ref{tab:exinc} and \ref{tab:examb} below illustrate interaction examples from the ShARC dataset between a human agent and an LLM. The Table \ref{tab:exinc} highlights an example of an incomplete question. It shows how the model requests for clarification due to lack of specificity in the question to provide a precise response.
The Table \ref{tab:examb} illustrates an ambiguous question. It shows how the model initially provides an incorrect response, and corrects it based on the human agent's input.
Table \ref{tab:turn} depicts an example of an interaction from AmbigNQ dataset in which the incorrect output of the single turn is answered correctly when LLM receives the additional context.

\begin{table}[H]
\centering
\setlength{\tabcolsep}{8pt}
\caption{Snippet from ShARC dataset: Incomplete Question. The highlighted span in Question represents the source of Incompleteness.}
\begin{tabular}{|p{1.7cm}|p{2cm}|p{3.5cm}|}
\hline
\multirow{3}{*}{\textbf{Data sample}} & \textbf{Question} & Does \hl{this country} have social security agreements with the UK? \\  \cline{2-3}
& \textbf{Context} & Other countries with UK benefits arrangements. The following countries have social security agreements with the UK: Kosovo, Mauritius, Montenegro, and New Zealand. \\ \cline{2-3}
 \hline
 \multirow{3}{*}{\textbf{Agent response}} & \textbf{LLM} & Which country are you referring to? \\ \cline{2-3}
 & \textbf{Human Agent} & Montenegro. \\ \cline{2-3}
 & \textbf{LLM (Revised)} & Yes. \\
\hline
\end{tabular}
\label{tab:exinc}
\end{table}

\begin{table}[H]
\sethlcolor{pink}
\centering
\setlength{\tabcolsep}{8pt}
\caption{Snippet from ShARC dataset: Ambiguous Question. The highlighted span in Context represents the source of ambiguity.}
\begin{tabular}{|p{1.7cm}|p{2cm}|p{3.5cm}|}
\hline
\multirow{3}{*}{\textbf{Data sample}} & \textbf{Question} & Can I get Housing Benefit? \\  \cline{2-3}
& \textbf{Context} & \hl{If you're single and under 35}, you can only get Housing Benefit for bed-sit accommodation or a single room in shared accommodation. \\ \cline{2-3}
 \hline
 \multirow{3}{*}{\textbf{Agent response}} & \textbf{LLM} & Yes, if you're single and under 35, you can get Housing Benefit for bed-sit accommodation or a single room in shared accommodation. \\ \cline{2-3}
 & \textbf{Human Agent} & No, I am not single and under 35. \\ \cline{2-3}
 & \textbf{LLM (Revised)} & No. \\
\hline
\end{tabular}
\label{tab:examb}
\end{table}

\begin{table}[H]
\sethlcolor{green}
\centering
\setlength{\tabcolsep}{8pt}
\caption{Snippet from AmbigNQ dataset: The highlighted span in the Human Agent response shows the turnwise additional context, which makes the LLM answer accurate.}
\begin{tabular}{|p{1.7cm}|p{2cm}|p{3.5cm}|}
\hline
\multirow{3}{*}{\textbf{Data sample}} & \textbf{Question} & Where was the first known government in the Western world to have a representative government? \\  \cline{2-3}
& \textbf{Ground Truth} & Roman Republic. \\ \cline{2-3}
\hline
\multirow{3}{*}{\textbf{Agent response}} & \textbf{LLM Turn 1} & The first known government in the western world to have a representative government was Ancient Greece. \\ \cline{2-3}
& \textbf{Human Agent} & It \hl{began in 509 BC}, so which government was it. \\ \cline{2-3}
& \textbf{LLM Turn 2} & The first known government in the western world that began in 509 BC was the Roman Republic.\\ \cline{2-3}
\hline
\end{tabular}
\label{tab:turn}
\end{table}

\end{document}